\documentclass[11pt]{article}

\usepackage{dialogue}

\usepackage{ifxetex}
\ifxetex
    \usepackage{fontspec}
    \setromanfont{Times New Roman}
\else
  \usepackage{cmap}
  \usepackage[T2A,T1]{fontenc}
  \usepackage[utf8]{inputenc}
  \usepackage{times}
  \usepackage{latexsym}
  \usepackage{substitutefont}
  \substitutefont{T2A}{\familydefault}{cmr}
\fi

\usepackage[russian,british]{babel}
\usepackage{url}
\usepackage{pgf}

\usepackage{covington} 

\usepackage{hyperref}

\dialogfinalcopy 

\title{RuSentNE-2023: Evaluating Entity-Oriented Sentiment Analysis \\on Russian News Texts}

\author{Anton Golubev \\
  Lomonosov Moscow  \\
  State University  \\
  {\tt antongolubev5@yandex.ru} \\\And
  Nicolay Rusnachenko \\
  Bauman Moscow State \\
  Technical University  \\
  {\tt rusnicolay@gmail.com} \\ \And
  Natalia Loukachevitch \\
  Lomonosov Moscow  \\
   State University  \\
  {\tt louk\_nat@mail.ru} \\ 
  }

\date{}

\begin{document}
\maketitle
\begin{abstract}
The paper describes the RuSentNE-2023 evaluation devoted to targeted sentiment analysis in Russian news texts.  The task is to predict sentiment towards a named entity in a single sentence. The dataset for RuSentNE-2023  evaluation is based on the Russian news corpus RuSentNE having rich sentiment-related annotation. The corpus is annotated  with named entities and sentiments towards these entities, along with related effects and emotional states. The evaluation was organized using  the CodaLab competition framework. The main evaluation measure was macro-averaged measure of positive and negative classes.  The best results achieved were of 66\% Macro F-measure (Positive+Negative classes). 
We also tested ChatGPT on the test set from our evaluation and found that the zero-shot answers provided by ChatGPT reached 60\% of the  F-measure, which corresponds to 4th place in the evaluation. ChatGPT also provided detailed explanations of its conclusion. This can be considered as quite high for zero-shot application.

  
  \textbf{DOI:} 10.28995/2075-7182-2022-20-XX-XX
\end{abstract}

\selectlanguage{russian}
\begin{center}
  \russiantitle{RuSentNE-2023: aнализ тональности по отношению к сущностям \\ в русскоязычных новостных текстах}

  \medskip \setlength\tabcolsep{0.5cm}

  \vspace{1.0 cm}
  \begin{tabular}{ccc}
    \textbf{Голубев А.А.} & \textbf{Русначенко Н.Л.} & \textbf{Лукашевич Н.В.}
    \\
    МГУ им. Ломоносова & МГТУ  им. Баумана&    МГУ им. Ломоносова                                              \\
    Москва, Россия &  Москва, Россия &  Москва, Россия      \\               
    {\tt antongolubev5@yandex.ru} &  {\tt rusnicolay@gmail.com} & {\tt louk\_nat@mail.ru}
  \end{tabular}
  \medskip
\end{center}

\begin{abstract}
  {В статье описывается тестирование RuSentNE-2023, посвященное таргетированному анализу тональности в русскоязычных новостных текстах. Задача участников  состояла  в том, чтобы предсказать тональность по отношению к именованному объекту в предложении. Датасет тестирования RuSentNE-2023 основан на корпусе российских новостей RuSentNE, в котором размечено несколько типов являений, связанных с тональность.  Корпус аннотирован именованными сущностями,  тональностью по отношением к этим сущностям, размечены последствия для сущностей в связи описываемыми ситуациями и эмоциональные состояниями сущностей.Тестирование  было организовано на основе специализированного сайта для тестирований  CodaLab. Основной мерой оценки было макроусреднение положительных и отрицательных классов. Наилучший  результат, полученный участниками,  был  66\% макро-F-мере (положительные + отрицательные классы).
Мы также протестировали ChatGPT на тестовом наборе нашего тестирования и обнаружили, что zero-shot (без обучения) ответы ChatGPT достигают 60\% F-меры, что соответствует 4-му месту в тестировании RuSentNE. Модель ChatGPT также предоставила подробные пояснения к своему заключению. Этот результат можно считать достаточно высоким для применения в условиях zero-shot.}
  
  \textbf{Ключевые слова:} таргетированный анализ тональности, именованная сущность, новостные тексты, ChatGPT
\end{abstract}
\selectlanguage{british}

\section{Introduction}
\label{intro}

%
%
Sentiment analysis studies began  with the general task setting, in which the general sentiment of a sentence or a text should be detected. Currently, so-called targeted sentiment analysis is intensively discussed, in which a model should determine the attitude towards specific entities, their aspects (properties), or topics. 

Targeted sentiment analysis is especially important for news flow processing. It is assumed that news texts should be neutral, but in fact they contain a variety of information about positions on various issues of state bodies, companies, opinions of individuals, positive or negative attitudes of the mentioned subjects to each other. All these opinions are important for collecting and analysis. 

Currently, news sentiment seems understudied. For example, the search on paper titles in Google Scholar shows that number of papers devoted to sentiment in social media is three times larger than the number of paper discussing news sentiment. 

The specific features of news texts from the point of view of sentiment analysis are as follows \cite{loukachevitch2018extracting}:
\begin{itemize}
\item  these texts contain various opinions conveyed by different subjects, including
the author(s)’ attitudes, positions of cited sources, and relations of mentioned entities
to each other;
\item some sentences are full of named entities with different sentiments, which makes it difficult to determine sentiment for a specific named entity;
\item news texts contain numerous named entities with neutral sentiment, which means that the neutral class largely dominates;
\item significant share of sentiment in news texts  is implicit, for example can be inferred from some actions of entities.
\end{itemize}

The authors of \cite{hamborg2021towards} annotated a news corpus with sentiment and stated that sentiment in the news is less explicit, more dependent on context and the reader's position, and it requires a greater degree of interpretation. It was found that state-of-the-art approaches to targeted sentiment analysis perform worse on news articles than in other domains. The authors also point out that in 3\% of cases, the sentiment is depends on the position of the reader.

In this paper we present the Russian News corpus RuSentNE annotated with named entities and sentiment towards these entities, related effects and emotional states. We used the sentiment annotation of the RuSentNE corpus to organize the shared task RuSentNE-2023 within the framework of Dialogue evaluation series. 

\section{Related Work}
There were several international evaluations devoted to targeted sentiment analysis.

In 2012-2014 within the CLEF conference, RepLab events devoted to the evaluation of online reputation management systems were organized \cite{amigo2012overview,amigo2014overview}.
The task was to determine if the tweet content has positive or negative implications for the company’s reputation.

In the SemEval evaluation workshops 2015, 2016,  studies were devoted to aspect-based sentiment analysis (ABSA) in several domains. The task was to determine sentiment towards specific characteristics discussed in users' reviews such as food or service in restaurants \cite{pontiki2015semeval,pontiki2016semeval}.
In 2016-2017, topic-oriented sentiment analysis tasks were studied in the SemEval series of Twitter sentiment evaluations \cite{nakov2016semeval,rosenthal2017semeval}.

The latest trend in targeted sentiment analysis is the so-called Structured Sentiment Analysis, which involves extracting tuples from texts that describe opinions of the following form 
$\left<h, t, e, p\right>$, 
where $h$ is the opinion holder, $p$ represents sentiment (positive or negative), in relation to the entity $t$, expressed by means of the word or phrase $e$. Structured sentiment analysis can be divided into five subtasks: 
i) sentiment expression extraction, 
ii) sentiment object extraction, 
iii) sentiment subject extraction, 
iv) determination of the relationship between these elements, and 
v) sentiment extraction (positive or negative). 
Modern approaches aim to address these problems in a unified manner, generating the required tuples \cite{lin2022zhixiaobao}.
In 2022, the competition on structural sentiment analysis competition was organized as part of the SemEval-2022 international evaluation workshop \cite{barnes2022semeval}.

The relatively recent advent of \textit{transformers}~\cite{vaswani2017attention} cause a significant breakthrough in machine learning application across the variety of natural language processing tasks, including targeted sentiment analysis. Within the last five years, the significant amount of works were aimed on application of transformer components, namely the encoder~\cite{devlin2018bert} and decoder~\cite{alt_improving_2019}.  
In the case of the application of BERT, complemented by classification layer on top, has resulted in the appearance of a variety pretrained models employing different pretrain techniques~\cite{zhuang-etal-2021-robustly,alt2019improving}.  
Another group of authors studies conveying the structure of target in context, emphasizing the target in texts and forming a prompt-based input constructions~\cite{sun-etal-2019-utilizing,shin-etal-2020-autoprompt}.
To the best of our knowledge, the structured representation of the input contexts~\cite{morio-etal-2022-hitachi} as a part of sequence-to-sequence and graph-based models represents the latest advances in target oriented sentiment analysis.

For Russian, targeted sentiment evaluations were organized as a two-year series in 2015 and 2016.
In 2015, aspect-based sentiment evaluation in restaurant and car reviews was explored \cite{loukachevitch2015sentirueval}. 
During two years, methods for reputation monitoring tasks towards banks and telecommunication companies in tweets were evaluated \cite{loukachevitch2015entity}.  
The best classification results in 2015-2016 competitions were mainly based on the SVM method and the GRU neural network. 
Later, the results were significantly improved with the application of the BERT model \cite{golubev2020improving,golubev2021use}.

We see that most international and Russian evaluations are devoted to targeted sentiment analysis applied to social media and user reviews, not to news analysis. 
The general task of sentiment analysis devoted to classification of  Russian news quotes was studied in the ROMIP-2013 evaluation~\cite{chetviorkin2013evaluating}.

The work closest to ours is the NewsTSC corpus \cite{hamborg2021towards}. The authors selected articles of 14 US newspapers from the Common Crawl news crawl (CC-NEWS). Mentions of named entities such as PERSON or ORG were automatically identified, and corresponding sentences were extracted. The annotators should read the sentence and determine sentiment towards a highlighted named entity. In total, 3002 sentences were annotated, the neutral sentiment was most frequent (2087). Several BERT models were used in the experiments. The best results (58.8 macro F measure) were achieved  LCF-BERT \cite{zeng2019lcf}, where BERT was tuned on the Common  Crawl news corpus.

\section {RuSentNE Sentiment-Annotated Corpus}

The dataset RuSentNE-2023 constructed for the shared task, is based on the RuSentNE corpus, in which diverse sentiment-related phenomena are annotated. 

The  corpus RuSentNE includes news texts from  the NEREL dataset, annotated with 29 named entity types \cite{loukachevitch2021nerel}.   400 NEREL texts with the largest relative share of sentiment words according to the RuSentiLex lexicon \cite{loukachevitch2016creating} were selected for sentiment annotation in RuSentNE. 
The annotated named  entities were used as targets of sentiment in the RuSentNE corpus. 

Then sentiment-related tags  and relations towards NEREL entities were labeled in  the RuSentNE. The sentiment annotation includes 12  entity tags and 11 relation types, which describe sentiment, arguments, effects and emotions of entities mentioned in the text. 

\begin{figure}[h]
    \centering
    \includegraphics[width=0.70\textwidth]{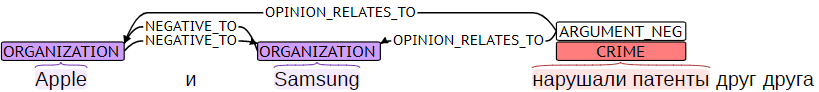}
    \caption{Example of sentiment annotation in sentence "Apple and Samsung infringed each other's patents" shows negative relations between companies and negative argument towards them}
    \label{fig:sentiment1}
\end{figure}

\begin{figure}[h]
    \centering
    \includegraphics[width=\textwidth]{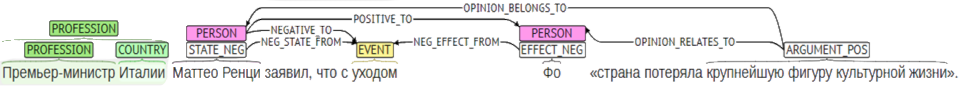}
    \caption{Example of sentiment annotation in sentence "Prime Minister of Italy Matteo~Renzi said that with Fo's departure, "the country has lost a major figure in cultural life."" demonstrates positive attitude from Matteo~Renzi towards Dario~Fo and also negative emotions stemminng from his death.}
    \label{fig:sentiment2}
\end{figure}

\begin{figure}[h]
    \centering
    \includegraphics[width=\textwidth]{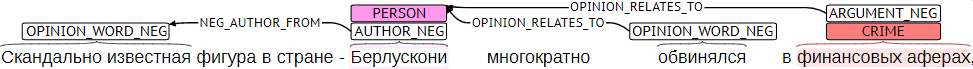}
    \caption{Example of sentiment annotation in sentence "Notorious figure in the country - Berlusconi has been repeatedly accused of financial fraud.""  shows negative opinion  of the author.}
    \label{fig:berlusconi}
\end{figure}

Figures~\ref{fig:sentiment1},  \ref{fig:sentiment2} and \ref{fig:berlusconi} show examples of annotation in the RuSentNE corpus.
In Figure~\ref{fig:sentiment1},
we can see that two companies are negative to each other (\texttt{NEGATIVE\_TO} relation); also  negative argument ("enfringed patents") is annotated that explains the attitude of two companies towards each other (\texttt{OPINION\_RELATES\_TO} relation). 

In Figure~\ref{fig:sentiment2}, we see that the Matteo~Renzi is positive to Fo (\texttt{POSTITVE\_TO} relation);  the explanation for this attitude  is given (\textit{major figure in cultural life}).The emotional state of Matteo~Renzi is negative because of the Fo death. Also negative effect for Dario~Fo is annotated, which originated from his death.   In Figure~\ref{fig:berlusconi} we see the author position towards Berlusconi depicted as tag \texttt{AUTHOR\_NEG}.

For the RuSentNE annotation, the BRAT annotation tool was used. The texts were annotated by both students and specialists in computational linguistics. 
All texts were then checked by a moderator, who discussed any identified issues' with annotators.

\section{RuSentNE-2023 Dataset}
\label{sec:dataset}

The annotation of the RuSentNE corpus was partially used for the RuSentNE-2023 evaluation.
Since RuSentNE contains 29 different classes of named entities, we selected  those entities that  are most frequent   objects of the targeted sentiment  in news texts. 
We selected the following  classes of entities for the evaluation:

\begin{itemize}
    \item PERSON — physical person regarded as an individual,
    \item ORGANIZATION — an organized group of people or company, 
    \item COUNTRY — a nation or a body of land with one government,
    \item PROFESSION — jobs, positions in various organizations, and professional titles,
    \item NATIONALITY — nouns denoting country citizens and adjectives corresponding to nations in contexts different from authority-related.
\end{itemize}

The distribution of entity types in the training, validation, and test sets is presented in Table~\ref{tab:entity_distribution}

\begin{table}[h]
    \centering
    \begin{tabular}{lcccccc}
    \hline
    Stage & Total & PERSON & ORG. & COUNTRY & PROF. & NATIONAL.\\
    \hline
    Train & 6637 & 1934 (29\%) & 1487 (23\%) & 1274 (19\%) & 1666 (25\%) & 276 (4\%)\\
    Development & 2845 & 857 (30\%) & 653 (23\%) & 686 (19\%) & 533 (24\%) & 116 (4\%)\\
    Final & 1947 & 480 (25\%) & 484 (25\%) & 363 (19\%) & 510 (26\%) & 110 (5\%)\\
    \hline
    \end{tabular}
    \caption{Distribution of  entity types in training, validation and test sets}
    \label{tab:entity_distribution}
\end{table}

For RuSentNE-2023 evaluation devoted to targeted sentiment analysis, we used a subset of annotated  entity tags and relations from the initial annotation of the RuSentNE corpus:

\begin{itemize}
    \item  \texttt{AUTHOR\_POS}, \texttt{AUTHOR\_NEG} tags describe the author's attitude  towards the tagged entity,
    \item  \texttt{POSITIVE\_TO}, \texttt{NEGATIVE\_TO} relations describe a relationship between two entities in a text,
    \item \texttt{OPINION\_RELATES\_TO} relation describes the relation from the opinion expressed in the text to the object of the opinion.
    \begin{itemize}
        \item \texttt{OPINION\_WORD\_NEG}, \texttt{ARGUMENT\_NEG} — negative to the object,
        \item \texttt{OPINION\_WORD\_POS}, \texttt{ARGUMENT\_POS} — positive to the object
    \end{itemize}
\end{itemize}
Entities without any sentiment annotations or relations directed towards them form the neutral class. 

In particular, the following examples of targeted sentiment in the NEREL-2023 dataset are obtained from the annotation presented in Figures~\ref{fig:sentiment1}, \ref{fig:sentiment2}, and \ref{fig:berlusconi} (Table \ref{tab:Examples}).

\begin{table}[h]
    \centering
    \begin{tabular}{lccl}
    \hline
    Source         & Entity& Sentiment &  Source in RuSentNE\\
    \hline
    
    Figure~\ref{fig:sentiment1}& Samsung & negative & \texttt{NEGATIVE\_TO}, \texttt{ARGUMENT\_NEG}\\
    Figure~\ref{fig:sentiment1}& Apple & negative & \texttt{NEGATIVE\_TO}, \texttt{ARGUMENT\_NEG}\\
    Figure~\ref{fig:sentiment2}& Fo & positive & \texttt{POSITIVE\_TO}, \texttt{ARGUMENT\_POS}\\
    Figure~\ref{fig:sentiment2}& Matteo~Renzi & neutral & absense of annotation\\
    Figure~\ref{fig:berlusconi}& Berlusconi & negative & \texttt{AUTHOR\_NEG}, \texttt{ARGUMENT\_NEG}\\
    \hline
    \end{tabular}
    \caption{Examples of sentiment labels in RuSentNE-2023 dataset}
    \label{tab:Examples}
\end{table}

In the original RuSentNE corpus, sentiment-related relations can be annotated across sentences.  In the RuSentNE-2023 evaluation, the targeted sentiment should be extracted from a single sentence. This implies that in a specific sentence, sentiment towards a target entity can be absent.
Therefore, it was then necessary to ensure that the source of sentiment  is located in the same sentence as the entity. If not, the sample was not included in the dataset. 
In the example below, the Tula transport prosecutor's office from the second sentence derives the positive annotation from the first one, which contradicts the relations from isolated second sentence:

\begin{itemize}
    \item \textit{<<The Tula transport prosecutor's office \textbf{defended the rights} of workers. The Tula transport prosecutor's office \textbf{filed two lawsuits against Russian Railways} for the recovery of child care allowances>>.}
\end{itemize}

The pre-trained RuCoreNews\textsubscript{small} spaCy model\footnote{\url{https://spacy.io/models/ru\#ru_core_news_sm}} was utilized to segment texts into sentences.

The initial RuSentNE corpus contains so-called nested named entities, that is one entity can be annotated within another entity. 
Only an  upper-level entity  was selected for the RuSentNE-2023 dataset. 
An example of nested entities can be seen in Figure \ref{fig:sentiment2}: entity \textit{Prime-minister of Italy} contains entity \textit{Italy}.

    
    
    

    

After the collection was formed, several post-processing steps were carried out. 
The same sentences that were included in the dataset multiple times according to different criteria were deduplicated. 
Examples with conflicting annotations were excluded. 
The minimum (40) and maximum (430) lengths of the text were chosen experimentally. 
Sentences excluded at this stage are presented below. 
The first sentence  contains conflicting annotation for Iran (Tel Aviv vs Syria, Hezbollah and Hamas),
while the following two sentences are too short:

\begin{itemize}

    \item \textit{<<Tel Aviv believes that the price of the withdrawal of Israeli troops from the Golan Heights should be political concessions from Syria - the weakening of its ties with its strategic ally \textbf{Iran} and the cessation of support for the Lebanese resistance movement Hezbollah and the Palestinian Hamas>>.}
    \item \textit{<<The \textbf{shepherd} was not injured>>.}
    \item \textit{<<\textbf{Pirates} fought back>>.}
    
\end{itemize}

Table~\ref{tab:distribution} presents the distribution of  sentiment scores in the training, validation and test sets.

\begin{table}[h]
    \centering
    \begin{tabular}{llcccc}
    \hline
    Type & Stage & Total & Positive & Negative  & Neutral\\
    \hline
    train & Train & 6637 & 856 (13\%) & 1007 (15\%) & 4774 (72\%)\\
    validation & Development & 2845 & 362 (13\%) & 438 (15\%) & 2045 (72\%)\\
    test & Final & 1947 & 269 (14\%) & 243 (12\%) & 1435 (74\%)\\
    \hline
    \end{tabular}
    \caption{Distribution of  sentiment scores in training, validation and test sets.}
    \label{tab:distribution}
\end{table}


We estimated the inter-annotated agreement in the RuSentNE-2023 dataset using duplicating sentences from the initial dataset. Cohen's Kappa is calculated as 0.5 (moderate agreement). But the Cohen's Kappa can be  unreliable in our case due to Kappa’s sensitivity to class imbalance \cite{feinstein1990high}. The percentage agreement between annotators is 84\%.

\section{Task Description}


In the RuSentNE-2023 competition, named entities  should be classified into one of three sentiment classes: positive, negative or neutral within the context of a single sentence. 
 Each sentence is annotated as follows:


\begin{itemize}
    \item  \texttt{entity} — object of sentiment analysis
    \item  \texttt{entity\_tag} — tag of this object (see Section~\ref{sec:dataset})
    \item  \texttt{entity\_pos\_start\_rel} — index of the initial character of the given entity
    \item  \texttt{entity\_pos\_end\_rel} — index of the next character after the last of the given entity
    \item  \texttt{label} — sentiment label 
\end{itemize}

Each entity has a three-scaled label. The following classes (labels) are used:
\begin{itemize}
    \item Negative (-1)
    \item Neutral (0)
    \item Positive (1)
\end{itemize}

 Participants are tasked with automatically annotating  each test sentence with the appropriate sentiment label for a specified entity.

As the primary evaluation metric, we adopt the F1-PN-macro, which averages the F1-measures of the positive and negative classes. Additionally, we calculated the traditional F1-macro measure as a supplementary metric.

\section{Results}

Over 15 participants took part in development stage of the competition. Table~\ref{tab:dev} illustrates the results obtained by the competitors during this preliminary  evaluation stage.

\begin{table}[!htp]
    \centering
    \begin{tabular}{lcc}
    \hline
    Participant         & F1-PN-macro (rank) & F1-PN0-macro (rank) \\
    \hline
    \textbf{mtsai}      & 70.94 (1)  & 77.63 (1)  \\
    \textbf{cookies}    & 69.89 (2)  & 76.74 (2)  \\
    \textbf{lsanochkin} & 68.11 (3)  & 75.36 (3)  \\
    Dmitry315           & 62.91 (4)  & 71.28 (4)  \\
    ryzhtus             & 62.34 (5)  & 70.57 (6)  \\
    mitrokosta          & 62.15 (6)  & 70.70 (5)  \\
    sag\_m              & 61.35 (7)  & 69.73 (7)  \\
    shershulya          & 60.07 (8)  & 69.26 (8)  \\
    s231644             & 59.99 (9)  & 69.10 (9)  \\
    antongolubev        & 57.73 (10) & 68.21 (10) \\
    ild                 & 57.46 (11) & 67.52 (11) \\
    GreatDispersion     & 57.00 (12) & 65.38 (12) \\
    abc111              & 55.25 (13) & 65.00 (13) \\
    baseline\_model     & 44.32 (14) & 57.89 (14) \\
    postoevie           & 41.09 (15) & 48.28 (16) \\
    angyling            & 35.37 (16) & 43.76 (17) \\
    AlexSMSU            & 31.94 (17) & 49.25 (15) \\
    \hline
    \end{tabular}
    \caption{Results of the Development evaluation stage,
    ordered by F1-PN-macro;
    participant nicknames with top-3 results are bolded;
    baseline\_model results correspond to the application of the baseline model}
    \label{tab:dev}
\end{table}

The results of the final evaluation stage are illustrated in Table~\ref{tab:final}.
Participants were not limited in the number o submissions due to the relatively high threshold provided for both evaluation stages with 1K submissions limit. 
During the final stage, those participants who did not exceed the baseline result chose not to include their submissions in the leaderboard. In the following, we provide brief descriptions of the methods adopted by participants during the final evaluation stage.

\textbf{baseline\_model.} DeepPavlov~\cite{burtsev-etal-2018-deeppavlov} pre-trained conversational RuBERT\textsubscript{base}\footnote{\url{https://docs.deeppavlov. ai/en/master/features/models/bert.html}} is used. 
The model is fine-tuned treating the targeted sentiment analysis task as a question-answering problem \cite{sun2019utilizing}.




\begin{table}[h]
    \centering
    \begin{tabular}{lcc}
    \hline
    Participant         & F1-PN-macro (rank) & F1-PN0-macro (rank) \\
    \hline
    \textbf{mtsai}      & 66.67 (1)   & 74.11 (2)    \\
    \textbf{cookies}    & 66.64 (2)   & 74.29 (1)    \\
    \textbf{lsanochkin} & 62.92 (3)   & 71.20 (3)    \\
    ChatGPT \texttt{**} & 60.06 (-)   & 70.79 (-)    \\
    antongolubev        & 59.64 (4)   & 69.04 (4)    \\
    sag\_m              & 59.33 (5)   & 68.71 (5)    \\
    mitrokosta          & 58.68 (6)   & 67.54 (6)    \\
    Dmitry315           & 53.60 (7)   & 62.92 (7)    \\
    ild \texttt{*}      & 53.20 (-)   & 63.78 (-)    \\
    abc111              & 49.98 (8)   & 61.32 (8)    \\
    Naumov\_al          & 46.96 (9)   & 54.92 (10)   \\
    baseline\_model     & 40.92 (10)  & 56.71 (9)    \\
    \hline
    \end{tabular}
    \caption{Results of the Final evaluation stage,
    ordered by F1-PN-macro;
    participant nicknames with top-3 results are bolded;
    <<\texttt{*}>> corresponds to post-evaluation submissions, made rightafter the end of the final stage;
    baseline\_model results correspond to the application of the baseline model; 
    <<\texttt{**}>> represents zero-shot answers of ChatGPT application baseline}
    \label{tab:final}
\end{table}

\textbf{mtsai.} 
The participant  experimented with the following the following models:
RuRoBERTa\textsubscript{large}\footnote{
\url{https://huggingface.co/sberbank-ai/ruRoberta-large}
} , 
XLM-RoBERTa\textsubscript{large}\footnote{\url{https://huggingface.co/xlm-roberta-large}}, 
RemBERT\footnote{\url{https://huggingface.co/google/rembert}}~\cite{rembert}.
Two separate transformers-based models were adopted: one model was used for the original input, while the other was applied to the input  with masked entities. 
The participant utilized the predefined <<\texttt{[MASK]}>> token for entities, identified with the Named Entity Recognition (NER) approach.
Weights ranging from 0 to 1.0 were applied to the output, with the sentiment classes assigned a weight of 1.0 and the neutral class assigned a weight of 0.1.
The participant also implemented threshold for neutral class:
when the neutral class has the highest probability, but its
value below threshold, they select the most probable class among “positive” and “negative”.
Participant employed an ensembling technique, which was based on a five-fold split of the dataset from the development stage. This technique involved training different transformers on various splits of data. In total, five models were used to ensemble the output.

\textbf{cookies.} 
The participant experiment with the following set of language models: 
    language models RuBERT\textsubscript{base}, 
RuBERT\textsubscript{large}, 
RuRoBERTa\textsubscript{large}.
In terms of the input representation, various masks and markers for entity representation were considered.
During the development phase, participant concluded that RuRoBERTa\textsubscript{large} illustrates the highest results across the other models of experiment set,
as well as the absence of difference across different entity representation formats. 
For the final version of the input representation format, participant decided to mark the entity within the input sentence.  
The best submission for the development stage represents a system based on 
RuRoBERTa\textsubscript{large}, 
fine-tuned using the inverse probability weighting technique.
In the final evaluation stage, participant also employed ensemble learning and implemented automatic annotation of the development set.




 
\textbf{lsanochkin.} The participant used a language model  with a prompting technique according to the following assumption: passing entity ($e$) jointly with the sentence\footnote{like question answering or natural language inference prompt~\cite{sun-etal-2019-utilizing}} ($T$) in the following format:
     <<\texttt{[CLS]}~$T$~\texttt{[SEP]}~$e$>>.
The model was fine-tuned on the input text format with transfromer-based classification pipeline (fully-connected layer on the top of a transformer model, cross-entropy loss). 
In terms of the embedding pooling for the sentiment class identification, the participant consider <<\texttt{[CLS]}>> token      as a sentence representation.

\textbf{antongolubev.} The participant utilized the DeepPavlov RuBERT-base model fine-tuned in natural language inference (NLI) problem setting together with transfer learning approach. The model was pre-trained on targeted sentiment analysis data from SentiRuEval~2015-2016 evaluations \cite{loukachevitch2015sentirueval} and automatically generated data from Russian news corpus \cite{golubev2021use}. In addition, the SMOTE~\cite{chawla2002smote} augmentation approach for increasing of positive and negative classes was held.

\textbf{sag\_m.} The participant's approach was based on the  text-to-text generation approach using the ruT5\textsubscript{large}\footnote{\url{https://huggingface.co/ai-forever/ruT5-large}} model. The generated output text is one of the possible sentiment labels for the analyzed named entity: "negative", "neutral" or "positive".
The application process involved experimenting with several variants of data preparation for the model input and applying output token filtering to derive the final class label. The best results were achieved using additional data for training: the CABSAR dataset\footnote{\url{https://github.com/sag111/CABSAR}}.

\textbf{mitrokosta.} The participant applied the RuBERT model. Entity token hidden states of the last layer were pooled.
In addition, the hidden states of all layers were processed by convolutions and those that relate to the entity were averaged again. The latter yelded some improvements in quality.

\textbf{ild.} 
The participant adopts SBERT\textsubscript{large} language model, 
pre-trained initially on financial news dataset\footnote{\url{https://huggingface.co/chrommium/sbert_large-finetuned-sent_in_news_sents}}.

\textbf{Dmitry315.} The participant used the RuBERT model for fine-tuning on the provided data, with pooling for embeddings related to the entity.

\textbf{Naumov\_al.}
The participant considers  the  task as a multiclass-classification problem and  experimented with Interactive Attention Network (IAN)~\cite{ma2017interactive}. 
Numerous experiments were conducted using different word vector representations\footnote{ELMo, RuBERT\textsubscript{large} and \textbf{XLM-RoBERTa\textsubscript{large}} (the latter was used in final model), combined with the original IAN model}.
The participant experimented with several parameters including
hidden\_dim in the LSTM layers, the learning rate, and batch size.
Alongside the competition dataset, the participant used CABSAR corpus\footnote{\url{https://github.com/sag111/CABSAR}}, which contains Russian-language sentences from various sources: posts of the Live Journal social network, texts of the online news agency\footnote{\url{lenta.ru}}, and Twitter microblog posts.

We observe that in all approaches exceeding the baseline model results, participants used neural language models. 
Analyzing the results, it is worth to conclude the importance of the following  findings, related to language model application~\cite{vaswani2017attention}:
\begin{enumerate}
    \item a pretraining technique plays role: the contribution of RoBERTa results in higher performance rather than original BERT for the same sizes language models.
    \item scale of the models is another direction: larger size of the model likely results in a higher baseline in the case of the fixed batch size and training evaluation methodology across all the model's sizes under consideration.
    \item application of ensembling techniques allows gaining  higher prediction results (the two first approaches in the final leaderboard).
    \item task abstraction: this is a particular case when entities might be masked, marked or prompted which may assist with reaching the desired outcomes.
\end{enumerate}

\section{ChatGPT in RuSentNE-2023 evaluation}

In \cite{zhang2022would} authors report state-of-the-art results in the stance detection domain with zero-shot application of ChatGPT.  The latter became a source of our inspiration to contribute with the related findings by analyzing responses for RuSentNE-2023 dataset. 

We applied a  conversational system that uses the  GPT-3.5 model \cite{zhang2022would}~\footnote{\url{https://platform.openai.com/docs/models/gpt-3-5}},which comes with a pretrained state\footnote{\texttt{text-davinci-002-render-paid}, 2022/11/22} corresponding to InstructGPT model. Access to this model is provided by the paid subscription service,  ChatGPTPlus. The hidden state of the model was trained on text collections up to June~2021, and it supports   up to 4,097 tokens inputs. The model is non-deterministic, implying that identical inputs can result in different outputs. We used a default temperature parameter. The sentences from the dataset, translated using the \texttt{googletrans} library\footnote{\url{https://pypi.org/project/googletrans/}}, were input into the ChatGPT model and were also incorporated into a prompt formatted as follows:

\begin{center}
    \textit{<<What is the attitude of the sentence \textbf{[translated sentence]} to the target \textbf{[translated entity]}? Select one from <<favor, against or neutral>> and explain why>>.}
\end{center}

Regular expressions with later manual validation were used to analyse model responses. For several examples ($<0.1\%$ of data), the entity was translated incorrectly, which led to the fact that the model denied entity to be in the sentence. In such cases, the neutral label was put. Due to the limit on the number of requests to ChatGPT model per hour, a shell script was written to send examples at a given frequency. The whole test  dataset was processed in 55 hours. 

The results of ChatGPT are included in Table~\ref{tab:final}. We can see that the model applied in zero-shot manner (without fine-tuning) took the fourth place in the evaluation reaching more than 60\%  F1-PN-macro.

\section{Analysis of Examples}
To analyze difficult cases, we extracted examples from the test set, which were erroneously classified by all (9) or almost all (8) participants. We encountered the following main problem cases (the examples below are translated from Russian): 

\textbf{1. Models fail to  distinguish between the subject and the  object of opinion}. For example, in the sentence "In 2011, Azerbaijan will increase pressure on international organizations in connection with the Nagorno-Karabakh problem", eight models inaccurately predicted negative sentiment towards Azerbaijan. However, according to the ChatGPT explanation, which seems correct in this context: "The sentence simply states a fact about the intentions of the Azerbaijani government without expressing any positive or negative opinion about it".

A similar issue can be seen in the sentence: "In it, Crowley announced that the Obama administration intends to continue to cooperate with the Israeli politician". Here most models inaccurately predicted positive sentiment towards  the Obama administration, while both the annotators and ChatGPT indicated neutral sentiment. 

\textbf{2. Models  do not distinguish  some sentiment adjectives} (evident for a human reader, but ambiguous) and fail to identify sentiment directed towards the target entity. For example, in sentence "In the very first days of Moiseev's stay at the clinic of JSC "Medicina", the patient was examined by the country's \textbf{leading} neurosurgeon, who did not find
any indication for surgery", nearly all models, including ChatGPT, failed to detect positive sentiment towards "neurosurgeon".

\textbf{3. Models fail to identify  sentiment that requires an understanding of the relationships between entities.} For example, in sentence "On the first day of 2011, numerous messages appeared on the \textbf{Microsoft} forum from Hotmail users complaining about the disappearance of all read   messages stored in their mailboxes". In this case, all models (including ChatGPT) did not detect the  negative sentiment towards Microsoft, the owner of Hotmail service.

\textbf{4. Models do not distinguish implicit sentiment}. For example, in sentence "Over the years of her film career, she has appeared \textbf{in more than 30 films and television series}, including the sci-fi film Forbidden Planet and the detective series "Honey West", for which the actress \textbf{received a Golden Globe film award} and was \textbf{nominated for an Emmy Award.}" both the models and  ChatGPT were unable to detect the positive sentiment toward the  actress.

We also encountered several issues with human  annotation within our dataset, particularly in sentences relating to  sporting events. Some annotators labeled negative relationships between competing athletes but it seems in sports such  relations should not be annotated. Additionally,  discrepancies arose among annotators concerning the annotation of  victories and defeats: while some treated such outcomes as factual information, others interpreted them as positive or negative sentiments towards the athletes. Similar variations were observed in the output from ChatGPT.


\section{Conclusion}
In this paper, we described  the RuSentNE-2023 evaluation devoted to targeted sentiment analysis in Russian news texts. We think that targeted sentiment analysis in news texts has not been thoroughly explored, as prior research primarily focuses on sentiment within user reviews and social media content. The distinct characteristics of news texts include a significant proportion of neutral named entities, a substantial presence of implicit sentiment, and the occurrence of multiple named entities with varying sentiments within the same sentence.

The objective of the RuSentNE-2023 evaluation was to predict sentiment towards a named entity within a single sentence, effectively classifying the sentiment into one of three categories: positive, negative, or neutral. The dataset used for the RuSentNE-2023 evaluation is derived from the Russian News corpus, RuSentNE, which contains a comprehensive sentiment-related annotation. The corpus is marked with named entities and the sentiment directed towards these entities, alongside associated effects and emotional states.

The evaluation was organized using  the CodaLab competition framework. The main evaluation measure was macro-averaged measure of positive and negative classes. 
There were 15 participants in the development stage of the evaluation, with 9 participants presenting their results in the final stage.
 The best results reached of 66\% Macro F-measure for the positive and negative classes. 

We additionally conducted experiments with ChatGPT on the test set of our evaluation, discovering  that the zero-shot responses of ChatGPT reached 60\% of the  F-measure, which corresponds to the fourth-place in the evaluation. ChatGPT also provided comprehensive explanations of its conclusions. The outcomes are considerably notable, given that this was a zero-shot application.

Our future plans involve improving the annotation of the RuSentNE corpus, taking into account the results of the participants and the explanations provided by ChatGPT. Subsequently, we aim to conduct an evaluation focused on structured sentiment analysis, specifically tailored for the extraction of four-tuples (the subject of sentiment, object of sentiment, sentiment, and sentiment expression). Furthermore, it is important to  eliminate sentence restrictions and advance towards classifying sentiment across sentences.








\section*{Acknowledgements}

The work was supported by the Russian Science Foundation
under Agreement No. 21-71-30003.
\bibliography{dialogue.bib}
\bibliographystyle{dialogue}



\end{document}